\documentclass[10pt,conference,a4paper]{IEEEtran}
\usepackage{times,amsmath,epsfig}
\usepackage{epstopdf}
\title{Brain-Inspired Inference on Missing Video Sequence}
\author{%
{}
\vspace{1.6mm}\\
\fontsize{10}{10}\selectfont\itshape
\,\\
\\
\fontsize{12}{12}\selectfont\ttfamily\upshape
\,Weimian Li, Baoyang Chen, Wenmin Wang\\
\,School of Electronics and Computer Engineering, Peking University

\vspace{1.2mm}\\
\fontsize{10}{10}\selectfont\rmfamily\itshape
\,\\
\\

\fontsize{9}{9}\selectfont\ttfamily\upshape
\,
}
\begin{document}
\maketitle



\begin{abstract}
In this paper, we propose a novel end-to-end architecture that could generate a variety of plausible video sequences correlating two given discontinuous frames. Our work is inspired by the human ability of inference. Specifically, given two static images, human are capable of inferring what might happen in between as well as present diverse versions of their inference. We firstly train our model to learn the transformation to understand the movement trends within given frames. For the sake of imitating the inference of human, we introduce a latent variable sampled from Gaussian distribution. By means of integrating different latent variables with learned transformation features, the model could learn more various possible motion modes. Then applying these motion modes on the original frame, we could acquire various corresponding intermediate video sequence. Moreover, the framework is trained in adversarial fashion with unsupervised learning. Evaluating on the Moving Mnist dataset and the 2D Shape dataset, we show that our model is capable of imitating the human inference to some extent.
\\[1\baselineskip]
\end{abstract}

\begin{keywords}
Video Inference, Brain Inspired, Generation, Transformation, Adversarial Learning
\end{keywords}

\section{Introduction}

Given two discrete static video frames, human could infer what happens in between according to the context. More importantly, people are not going to be limited into only one possibility, and could present various versions of the inference.

Inspired by such interesting intuition, in this work, we propose a novel end-to-end framework that imitates the human capability of inference and learns to generate both diverse and plausible video sequence to correlate two given discontinuous frames. Note that the task of video inference is quite different from the video interpolation. Interpolation needs to get generated intermediate frames as closer to the ground truth as possible, while inference focuses on the variety and rationality.

The study of video inference is still in its infancy, and there is few related work as well. However, our goal is similar to the task of video prediction, and thus we could borrow some ideas from it. Some work like Srivastava et al. \cite{srivastava2015unsupervised}, Mathieu et al. \cite{mathieu2015deep} and Kalchbrenner et al. \cite{kalchbrenner2016video} directly generate video in pixel level. In contrast, another category of works begin with learning the motion information within input video sequence, and then apply learned transform features on given frames to predict future frames \cite{van2017transformation} \cite{xue2016visual} \cite{oh2015action} \cite{finn2016unsupervised}. Generally, motion only occurs in a fraction of region of the scene while rest keeps invariant, so that prediction on all pixels is redundant and may accumulate more error. Therefore, in our framework, we employ latter transform-based method to generate video. Moreover, to extract transform features, we formulate our model in an encoder-decoder manner like most video prediction approaches do.

\begin{figure}
	\includegraphics[width=3in]{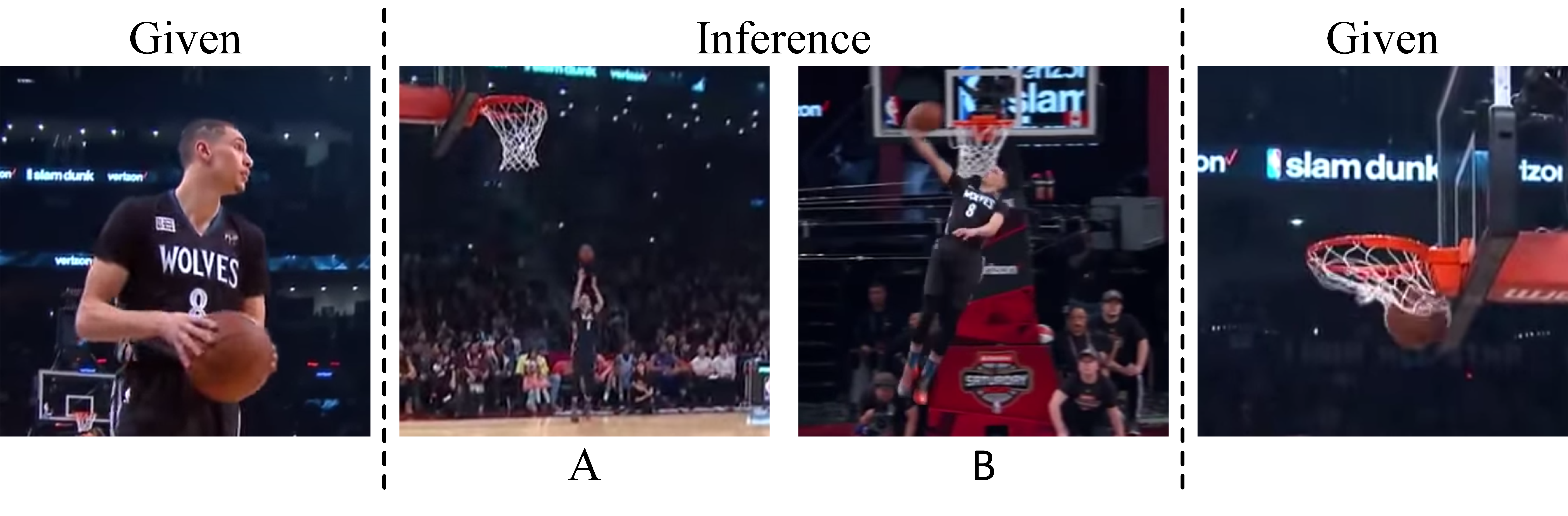}
	\caption{An example of human capability of inference. The first image shows
		a player holding a basketball, and the last one displays the basketball is in.
		Based on these two scenes, generally people could infer how he scores: may
		be shooting (A) or dunking (B). }
	\label{fig:illu}
\end{figure}

In order to imitate the variety and certain randomness of inference of human, we introduce a latent variable drawn from Gaussian distribution. Then via integrating different latent variables with learned transform features, the model could acquire more possible motion modes. Please note that these extended motion modes are not unreasonable. The initial transformation is learned from the actual input, and thus the variation that the latent variable brings are well-founded. It resembles a sudden inspiration flashes upon your mind changing your fact-based judgement of things, and we prefer to consider such variable as the imitation of inspiration.

\begin{figure*}[h]
	\includegraphics[width=7in]{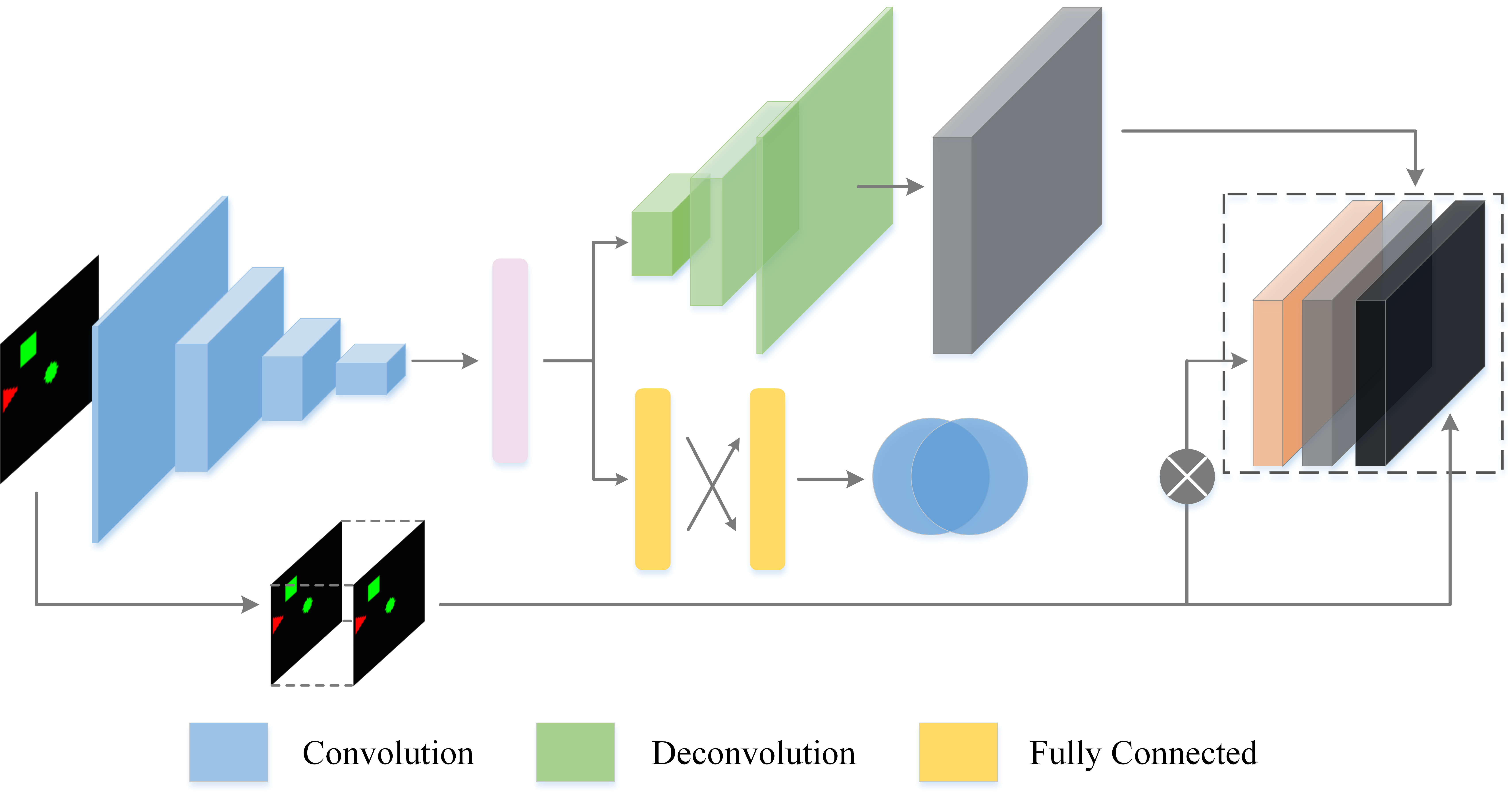}
	\caption{Pipeline of video imagination from single image. \textrm{In our framework, to produce one imaginary video, the input image is first encoded into a condition code and sent to transformation generator together with a latent variable. The generated transformation sequence is applied to input image later in volumetric merge network where frames are reconstructed with transformed images and volumetric kernels. Those four frames form one imaginary video. By sampling different latent variable from guassian distribution, our framework can produce diverse imaginary videos.} }
	\label{fig:model}
\end{figure*}

Inference does not pursue exactly the same with actual videos, and allows certain deviation as well as extension. As long as what we generate appear to be real and plausible, they could be considered acceptable inference. Therefore, traditional supervised learning used in video prediction is incompetent here. Instead, we train our model in adversarial fashion \cite{goodfellow2014generative}, which has achieved great success in unsupervised image generation \cite{zhao2016energy} \cite{nowozin2016f} \cite{uehara2016b}. Specifically, we put both the generated and real videos into a discriminator that would distinguish counterfeits from the real ones. Due to such training manner, our completed video sequences are going to gradually appears to be authentic. As for the design of discriminator, we employ the strategy in  Wasserstein GAN \cite{arjovsky2017wasserstein}, which is one of the most outstanding generative adversarial network so far.

We evaluate our design on two datasets, the Moving Mnist dataset \cite{srivastava2015unsupervised} and the 2D Shape dataset \cite{xue2016visual}. Conditioned on two discontinuous static frames,we are capable of generating a variety of reasonable short videos to correlate given images (see Fig), which shows that our model could imitate the human capacity of inference to some extent.

\section{Approach}

Inference on missing video sequence requires the capability of both understanding scenario and giving reasonable proposals. We take this challenge task as a generative problem with constraint. Fig.\ref{fig:model} shows the overall architecture of our framework.

our framework employs adversarial training and hence contains two part: generation part and discrimination part. 
In generation part, we first design a Scenario Encoder to understand start frame and end frame; 
Secondly, we use a Transformation-based strategy to generate transform features and masks. 
Moreover, we introduce an inspiration code to enable diverse inference. 
Finally, transform features and masks are applied to the start frame to produce inference frames. 
In discrimination part, we do not involve any supervised criterion, instead we design a spatio-temporal network to guarantee the completion video sequences are plausible.

\subsection{Scenario Encoder}
The information of the raw input, pairwise frames $<f_{start},f_{end}>$ indicating the start and the end frame in a video sequence, is limited. $<f_{start},f_{end}>$ depict a scenario representing both the appearance and dynamic change. A reasonable completion sequence needs to maintain the appearance as well as keep dynamic change consistent. Appearance information can be extract from frames while the difference image $f_{-}$ between $f_{start}$ and $f_{end}$ is a reasonable description of the dynamic change. We concatenate the first frame $f_{start}$ and the difference image $f_{-}$ in depth to serve as a 6-channel input of our framework. The input is then encoded into a scenario code.

Like inspiration is important in human inference, we believe an inspiration code is the key to make multiple inference. Before the scenario code sent to generator, it is concatenated with an inspiration code sampled from gussian distribution.
\subsection{Transformation-based Generator}
In order to better utilizing limited information, we build a transformation-based frame generator. Instead of reconstruct pixels from scratch, we use transform features to model motions, and synthesize frame through a mask. The transform features depict "how" object move and masks indicate "where". 

We employ affine transformation as the transform feature, because \cite{wang1993layered} has shown affine transformation can serve as a layered representation for motion analysis. One simple affine transform feature contains 6 parameters that can be formed as a matrix $T$. To transform image $I$ into a transformed image $\hat I$, firstly, $T$ determines a warping grids represent the coordinate correspondence between $I$ and $\hat I$. 
\begin{equation}
\begin{pmatrix}
\hat x_k \\
\hat y_k
\end{pmatrix} = T \begin{pmatrix}
x_k\\
y_k\\
1
\end{pmatrix}
\end{equation}

Each pixel value $\hat I(x_k,y_k)$ in $\hat I$ is produced through bilinear sampling.
\begin{align}
& \hat I(x_k,y_k) = \notag  \sum_i^W \sum_j^H I(i,j)max(0,1-\mid \hat x_k -i \mid)\\
& \hspace{15pt} max(0,1-\mid \hat y_k -j \mid)
\end{align}
To model more complex motions, multiple transform features $T^p$ are generated and applied to the start frame $f_0$ to form multiple transformed images $\hat f^p$s. Those $\hat f^p$s are composited into one inference frame $\hat f$ through point-wise multiply with a mask. Inference frames $\hat f$s are concatenated with $f_{start}$ and $f_{end}$ to form completion video sequence $\hat V$

\subsection{Adversarial training}
The task of inference means there is no precise reference or ground-truth, hence the traditional criterion like MSE is no longer appropriate. Recently adversarial training has been proved to have great performance in generative model. Especially Wasserstein GAN \cite{arjovsky2017wasserstein} accelerate and stabilize the training procedure. We design a spatial-temporal discriminator $D$ that give [judgement换] based on both spatial performance and temporal consistent. Let $P(V)$ represent the distribution of realistic video sequence and $P_T(V)$ denote generated video sequence. The generator loss $loss_g$ is defined as: 
\begin{equation}
loss_g =  -{E}_{v \sim P_T(V)}  D(v)
\end{equation}
The discriminate loss $loss_d$ is defined as: 
\begin{equation}
loss_d =  E_{v \sim P_T(V)}  D(v) - E_{v \sim P(V)}  D(v)
\end{equation}

Alternatively, we minimize the loss $loss_g$ once after minimizing the loss $loss_d$ 5 times until a fixed number of iterations.

\subsection{Implementation Details}
The missing video frames are generated recursively. Each time with two nonadjacent frames $<f_{t_1}, f_{t_2}>$ as input, the frame right in the middle $f_{\frac{t_2-t_1}{2}}$ is generated.

\section{Experiment}
\subsection{Implement Details}
In our experiments, the scenario encoder consists of 4
convolutional layers, while the decoder for generating masks
is formulated as a 3-layer deconvolutional network. Moreover,
we use a 2-layers fully connected network to generate transformation. The random variable sampled from the Guassian
distribution has 100 dimensions, and the scenario features
extracted by the encoder has 512 dimensions.

\subsection{Moving MNIST}
The moving MNIST dataset [1] consists of videos showing 2
digits moving inside a 64 x 64 frame. This dataset is generated
on the fly where the digits are chosen randomly from the
training set of MNIST dataset as well as assigned random
initial location, velocity and direction. In our experiments,
we generated 64,000 training video clips and 320 testing
clips each of length 5 frames. Taking the first and the last
frames as input, our model is trained to infer middle frames.
Both the generated and actual frames will be taken into the
discriminator, which is aim at improving the quality of our
inference but not making generated videos closer to real ones.
\subsection{2D Shape}
Next we evaluated our framework on a synthetic RGB video
dataset, the 2D shapes dataset [5] which contains three types
of objects: circles, squares, and triangles. The circles always
move vertically, squares horizontally and triangles diagonally
with random velocity within [0,5]. Since this dataset originally
is used for the task of predicting next frame given the first
frame, it only contains image pairs that have 2 consecutive
frames. Therefore, we extend it on the fly to convert image
pairs into video clips that have 5 frames. The training set has
20,000 video clips, and there are 500 clips for testing. Training
in the same setting as experiments on the moving MNIST, the
model learns to infer what might happen in middle 3 frames.
\subsection{Experiment Results}
Quantitative evaluation of generative models is a difficult,
unsolved problem [13], and some works attempt to explore an
uniform assessment method such as the Mean-Squared Error
(MSE), the Structural Similarity Index Measure (SSIM) [14]
and the Peak Signal to Noise Ratio (PSNR) [2]. However,
these evaluation systems are mainly designed for the task
of prediction, and the ground truth is used as an important
criterion. Therefore, existing assessments are not suitable in
our case. Moreover, for all we know, there are no published
works so far attempting to study the task of video inference
we propose in this paper, so that we have few available related
works for comparison. Based on the above, we could not
provide the quantitative evaluation of our approach and the
comparison with other ones for the present. To overcome
these problems, our future work will focus on the research
of feasible evaluation approaches for the video inference task.


\bibliographystyle{IEEEtran}

\bibliography{IEEEabrv,IEEEexample}

\begin{thebibliography}{10}
\providecommand{\url}[1]{#1}
\csname url@samestyle\endcsname
\providecommand{\newblock}{\relax}
\providecommand{\bibinfo}[2]{#2}
\providecommand{\BIBentrySTDinterwordspacing}{\spaceskip=0pt\relax}
\providecommand{\BIBentryALTinterwordstretchfactor}{4}
\providecommand{\BIBentryALTinterwordspacing}{\spaceskip=\fontdimen2\font plus
\BIBentryALTinterwordstretchfactor\fontdimen3\font minus
  \fontdimen4\font\relax}
\providecommand{\BIBforeignlanguage}[2]{{%
\expandafter\ifx\csname l@#1\endcsname\relax
\typeout{** WARNING: IEEEtran.bst: No hyphenation pattern has been}%
\typeout{** loaded for the language `#1'. Using the pattern for}%
\typeout{** the default language instead.}%
\else
\language=\csname l@#1\endcsname
\fi
#2}}
\providecommand{\BIBdecl}{\relax}
\BIBdecl

\bibitem{srivastava2015unsupervised}
N.~Srivastava, E.~Mansimov, and R.~Salakhudinov, ``Unsupervised learning of
  video representations using lstms,'' in \emph{International Conference on
  Machine Learning}, 2015, pp. 843--852.

\bibitem{mathieu2015deep}
M.~Mathieu, C.~Couprie, and Y.~LeCun, ``Deep multi-scale video prediction
  beyond mean square error,'' \emph{arXiv preprint arXiv:1511.05440}, 2015.

\bibitem{kalchbrenner2016video}
N.~Kalchbrenner, A.~v.~d. Oord, K.~Simonyan, I.~Danihelka, O.~Vinyals,
  A.~Graves, and K.~Kavukcuoglu, ``Video pixel networks,'' \emph{arXiv preprint
  arXiv:1610.00527}, 2016.

\bibitem{van2017transformation}
J.~van Amersfoort, A.~Kannan, M.~Ranzato, A.~Szlam, D.~Tran, and S.~Chintala,
  ``Transformation-based models of video sequences,'' \emph{arXiv preprint
  arXiv:1701.08435}, 2017.

\bibitem{xue2016visual}
T.~Xue, J.~Wu, K.~Bouman, and B.~Freeman, ``Visual dynamics: Probabilistic
  future frame synthesis via cross convolutional networks,'' in \emph{Advances
  in Neural Information Processing Systems}, 2016, pp. 91--99.

\bibitem{oh2015action}
J.~Oh, X.~Guo, H.~Lee, R.~L. Lewis, and S.~Singh, ``Action-conditional video
  prediction using deep networks in atari games,'' in \emph{Advances in Neural
  Information Processing Systems}, 2015, pp. 2863--2871.

\bibitem{finn2016unsupervised}
C.~Finn, I.~Goodfellow, and S.~Levine, ``Unsupervised learning for physical
  interaction through video prediction,'' in \emph{Advances in Neural
  Information Processing Systems}, 2016, pp. 64--72.

\bibitem{goodfellow2014generative}
I.~Goodfellow, J.~Pouget-Abadie, M.~Mirza, B.~Xu, D.~Warde-Farley, S.~Ozair,
  A.~Courville, and Y.~Bengio, ``Generative adversarial nets,'' in
  \emph{Advances in neural information processing systems}, 2014, pp.
  2672--2680.

\bibitem{zhao2016energy}
J.~Zhao, M.~Mathieu, and Y.~LeCun, ``Energy-based generative adversarial
  network,'' \emph{arXiv preprint arXiv:1609.03126}, 2016.

\bibitem{nowozin2016f}
S.~Nowozin, B.~Cseke, and R.~Tomioka, ``f-gan: Training generative neural
  samplers using variational divergence minimization,'' in \emph{Advances in
  Neural Information Processing Systems}, 2016, pp. 271--279.

\bibitem{uehara2016b}
M.~Uehara, I.~Sato, M.~Suzuki, K.~Nakayama, and Y.~Matsuo, ``b-gan: Unified
  framework of generative adversarial networks,'' 2016.

\bibitem{arjovsky2017wasserstein}
M.~Arjovsky, S.~Chintala, and L.~Bottou, ``Wasserstein gan,'' \emph{arXiv
  preprint arXiv:1701.07875}, 2017.

\bibitem{wang1993layered}
J.~Y. Wang and E.~H. Adelson, ``Layered representation for motion analysis,''
  in \emph{Computer Vision and Pattern Recognition, 1993. Proceedings CVPR'93.,
  1993 IEEE Computer Society Conference on}.\hskip 1em plus 0.5em minus
  0.4em\relax IEEE, 1993, pp. 361--366.

\end{thebibliography}

\end{document}